%% file: main.tex
\documentclass{article}

\PassOptionsToPackage{numbers, compress, sort}{natbib}
\usepackage[dblblindworkshop,final]{neurips_2026}

\newif\ifsubmission
\submissionfalse
\submissiontrue

\ifsubmission
\newcommand{\borisnote}[1]{}
\newcommand{\borisinline}[1]{}
\else
\usepackage[textsize=small]{todonotes}
\newcommand{\borisnote}[1]{\todo[color=cyan,inline]{boris: #1}}
\newcommand{\borisinline}[1]{\textit{\textcolor{orange}{[boris]: #1}}}
\fi

\usepackage{multirow}
\usepackage{makecell}
\usepackage{pifont}
\usepackage{xcolor}

\definecolor{deepgreen}{HTML}{55A868}
\definecolor{deepred}{HTML}{C44E52}

\newcommand{\cmark}{\ding{51}}
\newcommand{\xmark}{\ding{55}}

\usepackage{caption}
\usepackage[frozencache]{minted}

\AddToHook{env/minted/begin}{\let\colorbox\nobgcolorbox}
\usepackage{amsthm}
\newtheorem{example}{Example}

\usepackage{wrapfig}
\usepackage{graphicx}

\usepackage{listings}
\usepackage{xcolor}
\lstdefinestyle{mypython}{
  language=Python,
  basicstyle=\ttfamily\footnotesize,
  keywordstyle=\color{blue!70!black}\bfseries,
  commentstyle=\color{green!40!black}\itshape,
  stringstyle=\color{orange!80!black},
  numbers=left, numberstyle=\tiny\color{gray}, numbersep=8pt,
  frame=single, rulecolor=\color{gray!40},
  breaklines=true, showstringspaces=false,
  columns=flexible, tabsize=2,
  xleftmargin=1.5em,
}
\usepackage[utf8]{inputenc} % allow utf-8 input
\usepackage[T1]{fontenc}    % use 8-bit T1 fonts
\usepackage{hyperref}       % hyperlinks
\usepackage{url}            % simple URL typesetting
\usepackage{booktabs}       % professional-quality tables
\usepackage{amsfonts}       % blackboard math symbols
\usepackage{nicefrac}       % compact symbols for 1/2, etc.
\usepackage{microtype}      % microtypography
\usepackage{xcolor}         % colors
\usepackage{cleveref} % BORIS -- added
\usepackage{doi}      % BORIS -- added
\usepackage{enumitem}

\usepackage[normalem]{ulem}   % \sout for strikethrough; normalem keeps \emph italic
\usepackage{xcolor}
\usepackage{etoolbox}

\newtoggle{cr}
\toggletrue{cr}      % camera-ready: only "text after"
\newcommand{\replaced}[2]{%
  \iftoggle{cr}%
    {#2}%
    {\textcolor{red}{\sout{#1}}\,\textcolor{green!50!black}{#2}}%
}

\crefname{appendix}{Appendix}{Appendices}
\Crefname{appendix}{Appendix}{Appendices}

\title{Should SLMs Be in the Driver's Seat?\\
       {\large A Study of Local Models as Interpreters in Cloud-Written Agent Programs}}
\hypersetup{pdftitle={Should SLMs be in the Driver's Seat? A Study of Local
  Models as Interpreters in Cloud-Written Agent Programs}}
\workshoptitle{\href{https://slmw2026.github.io/}{SLM-Agents: 1st Workshop on SLMs for Agentic Systems}}
\workshoptitle{\href{https://slmw2026.github.io/}{SLMs for Agentic Systems, Paris, France, 2026}}
\workshoptitle{\href{https://slmw2026.github.io/}{SLMs for Agentic Systems}}

\author{%
  \textbf{Boris Radovič}$^{1,2,*}$ \qquad
  \textbf{Marco Canini}$^{1}$ \qquad
  \textbf{Veljko Pejović}$^{2}$ \\%[0.13em]
  \normalfont $^{1}$King Abdullah University of Science and Technology\\%\qquad
  $^{2}$University of Ljubljana, Faculty of Computer and Information Science \\[0.03em]% , Faculty of Computer and Information Science
  \normalfont $^{*}$\texttt{boris.radovic@kaust.edu.sa}
}

\begin{document}

\maketitle

\vspace{-1em}
\begin{abstract}
The most accurate agents executing user-provided tasks are driven by large, cloud-hosted models, which solve them by reading the private data of the user. Agents powered by models small enough to run on edge devices keep that data local, but yield far worse results. A third option is to have cloud models write a program that devices execute: the data stays local, but as the program is written before any data is seen, it fails whenever a step depends on interpreting content that appears only at run time. We explore an intermediate architecture, in which the cloud-written program defers bounded questions over user data to a small local model through an ordinary function call. On AppWorld this improves the success rate of a code-only agent in all six combinations of backbone model and test partition we evaluate, by up to 14.9pp, recovering a substantial part of the gap to a cloud-hosted ReAct loop while keeping user data on the device.
\end{abstract}

\input{sections/intro}
\input{sections/rw}
\input{sections/methodology}
\input{sections/eval}
\input{sections/conclusion}

\begin{ack}
This work was in part supported by the Slovenian Research Agency through the ``approXimation for adaptable distributed artificial intelligence'' (N2-0393) project.
\end{ack}

\bibliographystyle{plainnat}
\bibliography{main}

\appendix
\crefalias{section}{appendix}
\crefalias{subsection}{appendix}

\input{appendix/rw}

\input{appendix/examples}
\input{appendix/reproducibility}

\end{document}

%% file: sections/intro.tex
\section{Introduction}
\label{sec:introduction}

Users increasingly treat their smartphones and other \replaced{nearby}{}edge devices as personal assistants~\cite{Trivedi2024-qn,Erdogan2024-bh,Chen2024-hf}, asking them to carry out a range of tasks and expecting these tasks to be carried out correctly, quickly, and efficiently. State-of-the-art methods address such requests with agents~\cite{Yao2022-pw,Wang2024-rm}, i.e., systems in which a language model (LM) interacts with the environment on behalf of the user. The LM acts as the ``engine'' of the agent, and choosing it involves a fundamental trade-off, since \emph{which} LM is used also determines \emph{where} such a model can be run. Large LMs (LLMs) obtain by far the best task success rates, but they are typically too large to fit in the memory of an edge device and/or too compute-intensive for such devices, and are thus served on the cloud. Small LMs (SLMs) can instead run locally on the device of the user, but, as is generally expected from the scaling behavior of LMs~\cite{Kaplan2020-es}, are considerably weaker, especially in multi-step tasks~\cite{Haque2025-bq}.

\setlength{\columnsep}{14pt}
\begin{wrapfigure}[15]{r}{0.464\textwidth}
\vspace{-\intextsep}
\centering
\includegraphics[width=\linewidth]{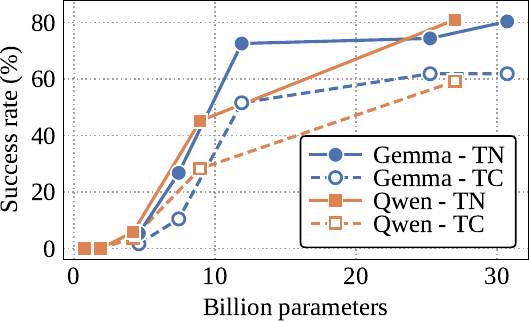}
\par\vspace{0.064pt}
\caption{Success rate of the ReAct agent on the AppWorld benchmark on the Test Normal (TN) and Test Challenge (TC) partitions, w.r.t. the number of LM parameters.}
\label{fig:react-wrt-size}
\end{wrapfigure}

\Cref{fig:react-wrt-size} quantifies the gap between SLMs and LLMs in agentic settings by running ReAct \cite{Yao2022-pw} agents on AppWorld~\cite{Trivedi2024-qn} with two LM families (Gemma 4, Qwen 3.5). Consistent with results reported by~\citet{Rainone2026-op}, the success of the agents degrades as the model size shrinks. No model below five billion parameters exceeds 5.9\% success rate on the Test Normal and 3.8\% on the Test Challenge partition, and the sub-two-billion LMs we evaluate solve no task at all. The sizes that make a local ReAct agent work reasonably well are also the ones that do not fit on edge devices: the $12$B Gemma model reaches 73\% success rate, but its weights alone require $\sim\!6$ GB in 4-bit precision, before the KV cache and before the memory the rest of the system needs. Even if memory were not an issue, running the ReAct loop locally would incur a substantial energy cost \cite{Wen2025-as}, which on battery-powered devices is undesirable. Based on these observations, we argue that at present SLMs are not ready to ``drive'' agentic tasks on their own.

At first glance this settles the matter in favor of the cloud, and indeed most deployed assistants are built this way. Such a choice, however, is problematic because of \emph{how} these agents consume data. In a ReAct-style loop, which is the dominant paradigm, the agent does not merely issue commands: rather, it inspects the outputs of the commands it issues and decides its next action based on what it has seen. For instance, when the agent issues a command to list the emails of the user, those emails are appended to the context and are processed in the next iteration. As ReAct agents issue short commands and constantly analyze their output, the result is a continuous flow of personal content leaving the device. Such a flow is problematic on several counts: it may violate the privacy of the user \cite{Haque2025-bq,Erdogan2024-bh}, it consumes bandwidth, and it is expensive, since providers bill by the number of input and output tokens~\cite{Narayan2025-xp,Kim2023-rh}. Lastly, relying entirely on the cloud leaves the local compute untapped, whereas exploiting this resource can reduce cost by up to 30$\times$~\cite{Narayan2025-xp}.

A less explored family of agents avoids this flow altogether. Instead of interleaving reasoning and observation, in a code-first agent the LLM writes a program for the user request\replaced{ and executes it}{, which then runs} end-to-end on the edge device where the data resides \cite{Trivedi2024-qn,Wen2025-as}. The cloud receives the user prompt and generates the script, but it never receives the data the program operates on. \replaced{As the program is written without observing user data, it can only do what plain code can do. }{}\replaced{Whenever a step towards solving the problem depends on properly interpreting content retrieved at run time, code alone will likely fail. This is because a program written before the data is read can only match patterns fixed in advance. In numerous situations, however, reliable pattern matching is difficult to pre-engineer.}{Thus, whenever a step towards solving the problem depends on properly interpreting content retrieved at run time, such an approach will likely fail: a program written before the data is read can only match patterns fixed in advance, and in numerous situations reliable patterns are difficult or impossible to pre-engineer.}

The limitations of the two paradigms are thus of a rather different nature. Cloud-based ReAct agents achieve high success rates, but sacrifice data locality, while a code-first approach keeps the data local but fails whenever a task requires data interpretation. This difference suggests dividing the work along a boundary that conveniently coincides with the privacy one. The structure of a task, i.e., which APIs to call, in which order, and under which conditions, typically contains no private data~\cite{Yu2026-ow} and is the part that genuinely benefits from a large model~\cite{Prasad2023-sv,Hou2024-fz}. The content the task operates on is private, but the reasoning it demands is often narrow~\cite{Narayan2025-xp,Belcak2025-uc}.

In this paper we explore this division: a cloud LLM writes a program for the user request and, wherever the program has to interpret user data, defers an ``atomic'' reasoning step\replaced{}{, i.e., a single, stateless question over program-supplied data,} to an SLM running locally. \replaced{We}{Similarly to \citet{debenedetti2026defeating}, we} instantiate the design as a single Python function exposed to the LLM, which turns a semantic step into an ordinary call returning a typed value, so that the SLM is only asked isolated questions and never asked to plan. On AppWorld \cite{Trivedi2024-qn}, this improves the success rate of a code-first agent in all six combinations of backbone LLM and test partition that we evaluate by between $2.2$pp and $14.9$pp, recovering a substantial part of the gap that separates code-first agents from ReAct.

%% file: sections/rw.tex
\section{Related Work}
\label{sec:related_work}

We summarize here the closest lines of work and defer the full discussion to \Cref{app:extended_related_work}.

\textbf{Programs instead of loops.} PAL~\cite{Gao2022-mj} and Program-of-Thoughts~\cite{Chen2022-mh} showed that offloading the deterministic part of a task to an interpreter is more reliable than carrying it out in tokens. \citet{Trivedi2024-qn} bring the idea to agentic tasks, replacing the ReAct loop~\cite{Yao2022-pw} with a single Python program. Such programs need not be pure code: ViperGPT and VisProg give the code-generating LLM API access to a set of vision models and let it compose their usage~\cite{Suris2023-yy,Gupta2022-ac}\replaced{. Both, however, are tied to a single domain and query those models in the cloud, so execution never becomes local.}{, and CaMeL lets the program call a ``quarantined'' LLM to parse untrusted data as a defense against prompt injection~\cite{debenedetti2026defeating}. All of them, however, run these models in the cloud, so execution never becomes local; ViperGPT and VisProg are moreover tied to a single domain.}

\textbf{\replaced{SLMs on the device}{SLM capabilities}.} SLM accuracy degrades as the context grows. This degradation is already pronounced when the SLM is only asked to select a tool for a single step of an agentic task~\cite{Zhang2026-pd,Paramanayakam2025-av}, let alone when it has to chain multiple such steps to reach a goal~\cite{Haque2025-bq}. Consequently, systems that do run complete agentic tasks with SLMs make restrictive assumptions -- TinyAgent, for instance, has a fixed set of only $16$ tools~\cite{Erdogan2024-bh}. A single question over a single bounded piece of data does not enter this regime. We thus confine the SLM to answering exactly such questions over the context the program hands it, which is precisely where small models have been shown to be reliable~\cite{Narayan2025-xp,Belcak2025-uc}.

% , and is precisely where SLMs are reliable~\cite{Narayan2025-xp,Belcak2025-uc}; this is the only role we assign to SLMs.

% Systems splitting the work across clients and servers exist: 
\textbf{Local-remote collaboration.} Systems that split agentic tasks between device and cloud differ in the role they assign to the device: SLMs may extract information from documents for a remote aggregator~\cite{Narayan2025-xp}, execute and verify the steps of a cloud-generated plan~\cite{Yi2026-wo}, or run code the cloud synthesizes from periodically uploaded sensor readings~\cite{Dehury2026-ye}; a separate line keeps the loop in the cloud and has a local model sanitize or re-write what is uploaded~\cite{Yu2026-ow,Yuan2026-kj,Siyan2024-zf}. All of them either target a narrow domain or keep the remote model involved while the task runs, and none quantifies how often a code-first agent fails simply because its program cannot interpret the data it retrieves.

%% file: sections/methodology.tex
\section{Cloud Planner with On-Device Interpreter for Edge Agentic Tasks}
We consider two actors: a compute-powerful cloud node running an LLM and a compute-constrained edge device, which can only run an SLM -- by SLM we indicate any LM that fits within the memory budget of the device, and by LLM any that does not. As there is a one-to-one mapping between models and nodes, we drop qualifiers and write LLM for the cloud model and SLM for the client model. The client receives agentic requests from the user and is tasked with carrying them out. In doing so, other than maximizing the accuracy, its desiderata are:
\begin{enumerate}[nosep, leftmargin=*, label=(\arabic*)]
\item minimize local compute, and thus energy consumption,
\item keep the data used during execution local, and
\item minimize LLM involvement.
\end{enumerate}
The server has no direct interest in the first two points. The third, however, can be seen as shared by the two actors, though for different reasons: the client wants to minimize the number of LLM tokens to reduce cost, while the server may be interested in minimizing the effort spent on a client so as to serve a larger number of clients concurrently. That is, with the same infrastructure, by minimizing the resources (time, tokens, compute, memory) allocated to a client, the server can serve more clients.

% With respect to the second desideratum, we assume that the client prompt does not contain private data and can be shared with the server. In the opposite case, the methodology is compatible with the mechanisms discussed by \citet{Siyan2024-zf} and \citet{Yu2026-ow}.

With respect to the second desideratum, while the data the task operates on should always remain confined to the device, we assume the prompt does not contain private information and can be shared with the server. Otherwise, our method is compatible with the prompt sanitization mechanisms~\cite{Siyan2024-zf}.

\subsection{Execution loop}

We study whether multi-step agentic tasks can be carried out by a script that runs end-to-end on the device: the LLM writes the program, and the SLM is queried only where a step requires interpreting data. Akin to the Full-Code (FC) agent~\cite{Trivedi2024-qn}, our framework proceeds as follows:
\begin{enumerate}[nosep, leftmargin=*, label=(\arabic*)]
    \item The edge device sends the user request to the LLM, which selects the APIs the program will use.
    \item The LLM, given the request, the documentation of the selected APIs, and instructions, generates a Python script. The instructions expose the framework-provided \texttt{slm} function and, since running the SLM is by far the dominant cost on the device, advise using it only where a code-only approach would be unreliable or impossible. The LLM ships the resulting program to the device.
    %The LLM, given the request, the documentation of the selected APIs, and instructions, generates a Python script. The instructions expose the framework-provided \verb|slm| function and, since running the SLM is expensive, advise the LLM to use it only where a code-only approach would be unreliable or impossible. The LLM ships the resulting program to the device;
    \item The device runs the program by involving the SLM at the points in which the LLM determined it to be necessary. If an exception is raised, the stack trace is sent back to the LLM, which reasons over it and generates another version, returning to step 2 with the error added to its context.
\end{enumerate}
The loop is capped at five attempts, after which the task is considered failed. If the initial API selection turns out to be incomplete, the LLM can extend it at run time without involving the device.

\begin{table}[b]
\centering
\caption{Qualitative comparison of the execution strategies.}
\label{tab:agents}

\begin{tabular}{r|ccccc}
\toprule
Agent & \makecell{Interprets\\run-time data} & \makecell{Local LM\\cost} & \makecell{User data\\to cloud} & \makecell{Cloud\\tokens} & \makecell{\replaced{Cloud\\invocations}{Cloud-device\\round trips}} \\
\midrule
Cloud ReAct & \cmark & none & every observation & high & one per step \\
Edge ReAct & \cmark & high & none & none & none \\
Full-Code (FC) & \xmark & none & request, traceback & moderate & $\leq 5$ \\
Ours & \cmark & low & request, traceback & moderate & $\leq 5$ \\
\bottomrule
\end{tabular}
\end{table}

% \Cref{tab:agents} positions the design against alternatives. Since execution is entirely local, only the user request and, upon failure, the exception may reach the cloud. On the device this costs a block of Python that calls APIs and performs no compute-intensive work, plus one SLM call per interpretation step over a bounded context; an all-local ReAct loop, by contrast, keeps the data local at the price of one LM invocation per step over a context that grows with every observation.

\Cref{tab:agents} positions our design: it interprets run-time data, which FC cannot; it invokes the local model only where needed, unlike local ReAct; and it uploads only the request and, on failure, the traceback.

\subsection{Programmatic interface to the SLM}
\label{sec:interface}

A challenge lies in how to make code interact with an SLM. Inspired by \citet{Suris2023-yy}, who provide a programmatic interface to models for solving visual question answering tasks, we define an ordinary Python function named \texttt{slm} and make it available by describing it in the instructions we give to the LLM\replaced{}{, much as CaMeL exposes its quarantined model~\cite{debenedetti2026defeating}}. Our goal is that, from the perspective of the program, the call \emph{behaves as closely as possible to an ordinary Python function}, by returning a value of a declared type or format.

\begin{wrapfigure}[8]{r}{0.43\textwidth}
\vspace{-1.2\intextsep}

\begin{minted}[frame=single, fontsize=\scriptsize,style=pastie]{python}
def slm(
    query: str,
    context: Any,
    context_description: str | None = None,
    keys: list[str] | None = None,
    choices: list[str] | None = None,
    output_type: str = "str",
    optional: bool = False,
) -> Any:
\end{minted}
\end{wrapfigure}

A call passes a question (\texttt{query}) and the data it ranges over (\texttt{context}). \replaced{The remaining arguments exist to restrict and guarantee the format of the response value}{The remaining arguments shape the input and constrain the output}: \texttt{choices} restricts the answer to a fixed vocabulary, \texttt{output\textunderscore type} declares the type into which the response is parsed and cast, \texttt{keys} projects away the fields of a JSON context that the question does not need,\replaced{ and}{} \texttt{context\textunderscore description} supplies format or ordering information that cannot be recovered from the serialized data alone\replaced{}{, and \texttt{optional} makes the function return \texttt{None} instead of raising when the answer is absent}. The function then performs all the overhead steps before and after the SLM invocation -- serialization, prompt construction, parsing, verification -- leaving the program with a plain function call, usable wherever the LLM deems some data interpretation necessary.

If the context does not contain the answer, the SLM is required to report its absence rather than fabricate a value. As a result, unless the \texttt{optional} argument is set, the call raises a \emph{not-found} exception~\cite{debenedetti2026defeating}, which is propagated to the LLM\replaced{}{ like any other exception}.

We present qualitative examples showcasing the usage of the \texttt{slm} function in \Cref{app:examples}.

%% file: sections/eval.tex
\section{Evaluation}
\label{sec:evaluation}

We evaluate the SLM agent on the AppWorld benchmark and its Test Normal (TN) and Test Challenge (TC) partitions, where the latter \replaced{represents}{contains} more complex tasks. We consider three LLMs, used for code generation by the SLM and FC agents and for driving the loop in the ReAct agent, specifically the Qwen 3.6 27B, the Qwen 3.5 122B-A10B, and the Qwen3-coder-next LLMs. For compactness, we refer to these models as Qwen-27B, Qwen-122B, and Qwen-coder, respectively. For the SLMs, we consider Qwen 3.5 models with sizes of up to 9B parameters\replaced{, all served in $8$-bit precision}{}. We report the success rate (SR), which indicates the percentage of correctly executed tasks. \replaced{To maximize reproducibility, we set the temperature to $0$ for all experiments.}{Further details about the experimental setup are provided in \Cref{app:reproducibility}.}

\subsection{How does the methodology compare to ReAct and FC agents?}

\begin{figure}[t]
  \begin{minipage}[t]{0.41\textwidth}
    \vspace{0pt}
    \centering
    \captionof{table}{SR for the considered agents and backbone LLMs and corresponding standard error, estimated with bootstrapping.}
    \label{tab:overall}
{\footnotesize
\renewcommand{\arraystretch}{1.11}    % shrinks every row's strut
\setlength{\tabcolsep}{4pt}
\begin{tabular}{ll|cc}
\toprule
LLM & Agent & TN & TC \\
\midrule
\multirow[c]{3}{*}{Qwen-coder} & ReAct & 59.5 {\tiny $\pm$ 3.7} & 29.5 {\tiny $\pm$ 2.2} \\
 & FC & 33.3 {\tiny $\pm$ 3.6} & 13.4 {\tiny $\pm$ 1.7} \\
 & SLM & 38.7 {\tiny $\pm$ 3.7} & 15.6 {\tiny $\pm$ 1.8} \\
\cline{1-4}
\multirow[c]{3}{*}{Qwen-122B} & ReAct & 57.1 {\tiny $\pm$ 3.9} & 38.6 {\tiny $\pm$ 2.4} \\
 & FC & 42.3 {\tiny $\pm$ 3.8} & 25.7 {\tiny $\pm$ 2.1} \\
 & SLM & 52.4 {\tiny $\pm$ 3.9} & 28.8 {\tiny $\pm$ 2.2} \\
\cline{1-4}
\multirow[c]{3}{*}{Qwen-27B} & ReAct & 81.0 {\tiny $\pm$ 3.0} & 59.2 {\tiny $\pm$ 2.4} \\
 & FC & 45.8 {\tiny $\pm$ 3.9} & 36.2 {\tiny $\pm$ 2.4} \\
 & SLM & 60.7 {\tiny $\pm$ 3.8} & 41.5 {\tiny $\pm$ 2.4} \\
\bottomrule
\end{tabular}
}
  \end{minipage}\hfill
  \begin{minipage}[t]{0.57\textwidth}
    \vspace{0pt}
    \centering
    \includegraphics[width=0.96\linewidth]{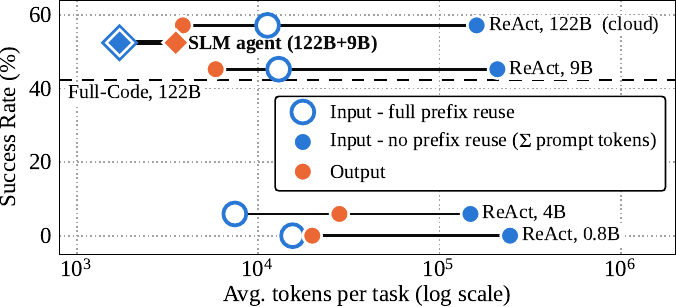}
    \caption{\replaced{(On-device) tokens vs. SR for Qwen-122B on TN. Each row of three markers is one agent, plotted against the avg. tokens its local model processes.}{SR vs. avg. tokens per task processed by the local model for Qwen-122B on TN. Each row represents one agent. For the SLM agent, only tokens of \texttt{slm} calls are counted. Cloud ReAct (122B, cloud-side tokens) and FC (dashed line, no local tokens) are shown for reference.}}
    \label{fig:sr-wrt-tokens}
  \end{minipage}
\end{figure}

We compare the agent we propose against the FC and ReAct agents and report the \replaced{corresponding}{}SR \replaced{}{of all three} in \Cref{tab:overall}. The results indicate that the SLM agent improves the performance of the FC agent on the TN partition by up to 14.9pp (in the case of Qwen-27B, from $45.8\%$ to $60.7\%$). In the case of the TC partition, on the other hand, the improvements are much more modest.

However, not every task requires an SLM during its execution, as evidenced by the non-zero SR of the FC agent. When analyzing the performance only on \replaced{the tasks in the TN partition,}{TN tasks} \replaced{which involve the SLM during execution}{in which the planner decided to use the \texttt{slm} function}, the SLM agent improves the SR of the FC agent from $0.0\rightarrow28.6$, $2.4\rightarrow38.1$, and $13.8\rightarrow48.3$ for the Qwen-coder, Qwen-122B, and Qwen-27B models, respectively.

\borisinline{We show two values for input tokens because the API endpoints of LLMs are stateless; if, after a tool output is re-directed to a different node, the KV-cache may need to be pre-filled again from scratch, increasing cost (\cite{Xu2023-xt})}

One of the driving motivations of the SLM agent is that having the local SLM answer only atomic questions is computationally lighter for the client than running an entire ReAct loop on-device. \Cref{fig:sr-wrt-tokens} shows this directly: with the same 9B model on the client, the SLM agent has a higher SR than a purely local ReAct agent (52.4\% vs 45.2\%) while the client processes on average between one and two orders of magnitude fewer input tokens -- the lower value under full prefix reuse -- and generates 1.8$\times$ fewer output tokens.\footnote{All Qwen SLMs run in thinking mode; token counts include reasoning tokens. Since \texttt{slm} returns only a scalar or short string, its output tokens are likely reducible via prompting or a non-reasoning model, though we have not evaluated this.} The saving is concentrated in input tokens because the ReAct agent must read the output of every command it issues, whereas the SLM only ever sees the values the cloud-generated program passes to it. A fully cloud-based ReAct execution reaches a somewhat higher success rate (57.1\%), but requires every token to be processed by the 122B model.

Because the SLM agent reduces the number of input and output tokens, its peak context is about $3\times$ smaller than that of an all-local ReAct agent using the same 9B model (median 3.7k vs 12.0k tokens). As a consequence, the SLM also requires less memory to keep the context.

By construction, the FC agent uses no tokens on the client. On the cloud, our instructions are a superset of the FC ones, adding\replaced{}{, per program,} \replaced{roughly }{}3.7k tokens describing \texttt{slm} and its best practices: our agent thus requires $\sim\!32\%$ more input tokens than FC, and slightly more output tokens, though never by more than 10\%. These extra instructions are also a confound, as the SLM agent improves on tasks that never invoke \texttt{slm} (e.g., +1.5pp for
Qwen-122B), a gain the mechanism cannot explain. This accounts for at most a fraction of the gains we report, leaving the majority attributable to the SLM.

\subsection{How large does the SLM need to be?}

\begin{figure}[t]
\begin{minipage}[t]{0.42\textwidth}
    \vspace{0pt}
    \centering
    \includegraphics[width=\linewidth]{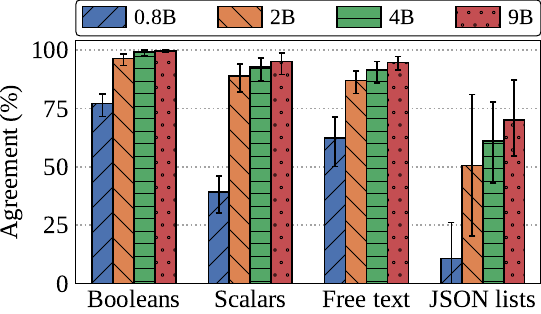}
    \caption{Agreement of Qwen SLMs with Qwen-9B; the 9B bar is self-agreement. Scalar types are \texttt{float}, \texttt{int}, and \texttt{item}.}
    \label{fig:agreement}
  \end{minipage}\hfill
  \begin{minipage}[t]{0.54\textwidth}
    \vspace{0pt}
    \centering
    \captionof{table}{Share of failing programs aggregated over TN, TC, and all LLMs, whose exception recurs in the agent's next program. $n$ counts failures that had a successor; each task's final program is excluded.}
    \label{tab:error-repeating}
{\footnotesize
\renewcommand{\arraystretch}{0.95}    % shrinks every row's strut
\setlength{\aboverulesep}{0.2ex}     % default 0.4ex
\setlength{\belowrulesep}{0.3ex}     % default 0.65ex
\begin{tabular}{lrrrr}
    \toprule
     & \multicolumn{2}{c}{FC} & \multicolumn{2}{c}{SLM} \\
    \cmidrule(lr){2-3}\cmidrule(lr){4-5}
    Failure type & $n$ & rep.\ (\%) & $n$ & rep.\ (\%) \\
    \midrule
    API error, 422 & 1070 & 35.6 & 966 & 34.3 \\
    Other API error & 205 & 16.1 & 174 & 15.5 \\
    API misuse & 310 & 8.1 & 299 & 10.7 \\
    Shape / attribute & 386 & 13.7 & 371 & 14.0 \\
    Safety guard & 76 & 10.5 & 56 & 10.7 \\
    % Other / environment & 24 & 45.8 & 33 & 51.5 \\
    \midrule
    SLM: info. not found & -- & -- & 448 & 55.6 \\
    \bottomrule
\end{tabular}
}
  \end{minipage}
\end{figure}

In the previous section we \replaced{reported results with a 9B model on the client}{considered a 9B client LM}\replaced{}{, which is already too large for many devices}. \Cref{fig:agreement} shows that the SLM need not be this large. We plot how often the value returned by \verb|slm| agrees with the one the 9B model produced on the identical request, replayed offline so that only the SLM varies. The 9B is a reasonable reference: an LLM-as-a-judge evaluation finds its answers correct about 95\% of the time on the TC partition. Decoding is not bit-reproducible across GPUs, so we also replay the 9B against its own trace; this self-comparison is the noise floor of the measurement.

For \replaced{structured}{constrained} output formats -- booleans, selections among candidate items, and numeric or short free-form strings -- a 2B model returns the identical value in 91.6\% of calls, 5.4pp below the 9B model's own reproducibility (97.0\%). Only at 0.8B is there a marked drop. Excluded from those numbers are the calls where the planner asks for a JSON object or list despite the \verb|slm| \replaced{contract}{instructions} forbidding it; agreement there is far lower, but so is the 9B's with itself (69.9\%). 

\begin{wraptable}[8]{r}{0.46\textwidth}
\vspace{-0.9\intextsep}
\centering
\footnotesize
\setlength{\tabcolsep}{4.5pt}
\caption{SR on TN w.r.t. SLM size.}
\label{tab:slm-size}
\begin{tabular}{l|ccc}
\toprule
SLM size & Qwen-coder & Qwen-122B & Qwen-27B \\
\midrule
None (FC) & $33.3$ {\tiny $\pm3.6$} & $42.3$ {\tiny $\pm3.8$} & $45.8$ {\tiny $\pm3.9$} \\
0.8B & $33.9$ {\tiny $\pm3.8$} & $48.8$ {\tiny $\pm3.9$} & $52.4$ {\tiny $\pm3.9$} \\
2B   & $36.9$ {\tiny $\pm3.7$} & $51.2$ {\tiny $\pm3.9$} & $61.3$ {\tiny $\pm3.7$} \\
% 4B   & X & Y & Z \\
9B   & $38.7$ {\tiny $\pm3.7$} & $52.4$ {\tiny $\pm3.9$} & $60.7$ {\tiny $\pm3.8$} \\
\bottomrule
\end{tabular}
\end{wraptable}
The above is an agreement study, not an end-to-end one, as the values \texttt{slm} returns re-enter the program and change what happens next. \Cref{tab:slm-size} reports the SR when repeating the evaluation end-to-end. \replaced{The 2B model causes only a small decrease compared to the 9B model}{The 2B and 9B LMs are within one standard error for all three planners}. The 0.8B model\replaced{}{, on the other hand,} loses more, but it still offers benefits over FC for Qwen-122B and Qwen-27B. Note that the sub-2B models solve no task at all when driving a ReAct loop (cf. \Cref{fig:react-wrt-size}), yet with the Qwen-27B planner they improve FC by 6.6pp (0.8B) and 15.5pp (2B), suggesting that \emph{answering questions over program-supplied data is better aligned with the current capabilities of SLMs than full planning}.

\subsection{Where does the gap to ReAct come from?}

Several factors contribute to the lower SR of the SLM agent relative to ReAct. We discuss the two we can measure. The first is that generated programs are not idempotent: attempts are not isolated, so a program that would be correct on the initial state can fail because an earlier attempt already modified that state, e.g., by sending a message. Re-executing the final program of each affected task in a fresh environment bounds this effect at $1.8$pp in any single configuration and under $0.5$pp on average, so cross-attempt state accumulation does not materially affect our results.

The second is that the planner is not told enough when an exception occurs. On failure, the stack trace is returned to the LLM, which generates the next program; for a not-found error the SLM agent additionally labels the cause as ambiguous question, wrong context, insufficient context, or unclassified. To test whether this is enough, we group failing programs by exception type and measure how often the next program the planner generates raises an exception of the same type; a recurrence indicates that the planner could not work out the cause from what it was told. \Cref{tab:error-repeating} shows that a not-found error recurs $55.6\%$ of the time, against \replaced{}{at most} $35.6\%$ \replaced{for the next most frequent type}{for any other type}. A recurrence does not lower SR by itself, but it consumes one of the five generation attempts and grows the context with a trace carrying nothing the planner did not already have. The two code agents agree within $3$pp on every exception type they share, so delegating to the SLM does not degrade ordinary recovery.

%% file: sections/conclusion.tex
\section{Conclusion and Future Work}
\label{sec:discussion_and_conclusion}

We studied a division of labor in which a cloud LLM writes the program for an agentic request and a local SLM is invoked from within it, only where a step requires interpreting \replaced{the}{user} data\replaced{ of the user}{}. This recovers much of what code-first agents lose by writing programs before observing any data, and, \replaced{against an all-local ReAct agent using the same model, it is both more accurate and substantially cheaper on the client}{given a capable planner, is both more accurate than a local ReAct agent using the same model and substantially cheaper on the client}. This paper reports an early instantiation of the design; below we discuss the open problems it surfaces and how we plan to address them\replaced{ in the framework we are building}{}.

\textbf{Privacy.} By construction, the only data that leaves the device is the user request and, on failure, the traceback, which the runtime emits and which may therefore embed user content. Analyzing the 2{,}917 exceptions raised across all our runs on the TC partition, we find such content -- messages, notes, or file contents -- in 12 of them ($0.4\%$). The majority of exceptions carry coarser information: timestamps echoed back when parsing fails, or negative facts such as ``file X does not exist''. Still, what leaves the device should be guaranteed rather than incidental. We thus plan to have the SLM sanitize the traceback locally, so that every upward payload is declared.

\textbf{Closing the gap.} The design does not yet match a cloud-hosted ReAct loop in terms of success rate. Part of the remaining gap comes from how little the planner learns when a program fails. All types of exceptions recur, but the effect is most pronounced when the SLM cannot find the requested information in the context (\Cref{tab:error-repeating}). We thus plan to have the SLM inspect the variables at the point when the exception happens and let it explain what went wrong, so that the planner can more easily generate a corrected program. Doing so safely is hard: such an explanation is derived from private content, so it must convey why the call failed without reproducing the data itself. \replaced{}{Alternatively, the SLM could attempt to recover from the failure on the device, where it has full access to the runtime state, avoiding the problem altogether, and escalate to the cloud only if local recovery fails. LMs can handle runtime errors~\cite{sun2026toward}, and weaker models can reuse programs written by stronger ones~\cite{cai2024large}; we thus also plan to study whether an SLM can recover from failures in the programs our planner writes.}

\textbf{Performance on edge devices.} Our resource claims rest on token and context measurements, which are only proxies for what matters on devices, e.g., latency and energy. Fewer tokens mean less compute and a smaller KV cache, so we expect the reductions we reported to carry over to those metrics. The actual savings, however, depend on hardware, which notoriously varies widely across edge devices. We plan to measure them once the method is developed into a full framework.

%% file: appendix/rw.tex
\section{Extended Related Work}
\label{app:extended_related_work}

\paragraph{ReAct, planning, programming.}
The predominant way of building agents is the ReAct loop, in which the agent reasons, emits an action -- expressed as text, JSON, or Python -- and observes its result, repeating until the task is completed~\cite{Yao2022-pw,Wang2024-rm}. Since larger models yield better agents, especially in multi-step tasks~\cite{Haque2025-bq}, such a loop is typically driven by an LLM served in the cloud, with the cost and privacy consequences discussed in \Cref{sec:introduction}. Running the loop on edge devices is impractical for the opposite reasons: accuracy is lower, the context grows with every observation and may overwhelm the client memory~\cite{Zhang2026-pd}. Also, involving the model at every action is compute-demanding~\cite{Wen2025-as}.

A separate line of research decomposes the task into smaller steps, either during execution when a step fails~\cite{Prasad2023-sv} or, more commonly, in an initial planning stage carried out before any data is observed, as in ReWOO~\cite{Xu2023-xt}. The plan may be a linear sequence, so that each step is handled by a ``more focused'' agent~\cite{Hou2024-fz,Chen2024-zs,Erdogan2025-xr,Wang2023-cx,Yihan2026-ya} or so that thinking can be cached and reused~\cite{Zhang2026-kp}, or a directed acyclic graph whose nodes are steps and whose edges are data dependencies~\cite{Erdogan2024-bh,Kim2023-rh,Shen2023-oi,Giusti2025-yo,Wei2025-wx}. Since nodes may be traditional functions, the LM need not be involved in every step, which makes these methods more efficient than ReAct in terms of tokens~\cite{Kim2023-rh,Xu2023-xt} and in some cases even more accurate~\cite{Xu2023-xt}. The relief is however partial: every node that involves language understanding is still a separate \emph{cloud} invocation over the data that node retrieves.

Such plans are, at the end of the day, no more than programs, i.e., defined sequences of steps to achieve a goal. PlanCompiler makes this explicit, as it ``compiles'' an LLM-generated plan into a Python script~\cite{Harikumar2026-rz}, while PAL and Program-of-Thoughts had already shown that offloading the deterministic part of a task to an interpreter is more reliable than carrying it out in tokens~\cite{Gao2022-mj,Chen2022-mh}. Closest to our approach, ViperGPT and VisProg give the code-generating LLM API access to a set of vision models and let it compose their usage in code~\cite{Suris2023-yy,Gupta2022-ac}, \replaced{}{CaMeL lets the program delegate the parsing of untrusted data to a quarantined LLM~\cite{debenedetti2026defeating,willison2023dual}, }while~\citet{Trivedi2024-qn} explore carrying out agentic tasks using Python code alone. \replaced{}{These methods, however, are either tied to a single domain~\cite{Suris2023-yy,Gupta2022-ac}, keep the cloud involved whenever data must be interpreted~\cite{debenedetti2026defeating}, or run programs written without observing any data, which fail whenever a step depends on content available only at run time~\cite{Trivedi2024-qn}. CaMeL, in particular, is methodologically similar to our approach: a ``privileged'' LLM writes code that may delegate the parsing of data to a ``quarantined'' LLM~\cite{debenedetti2026defeating}. The goal of CaMeL, however, is security, specifically preventing prompt injection, rather than data locality. While the authors note that the quarantined LLM could run locally, they evaluate the method only with cloud models; whether models small enough for edge devices can fill this role, and at what cost to the device, is the question we study.}

\paragraph{SLMs for tool calling and agentic tasks.}
A natural candidate to make raw data visible at run time without exposing it to the cloud is a local SLM. Turning an LM into an agent requires selecting a function and generating the arguments that carry out a task in the environment~\cite{Chen2024-sa,Schick2023-fy}, and several works study this task specifically on edge devices~\cite{Lin2024-ce,Chen2024-hf,Kadekodi2025-sq,Paprunia2025-cm}. Accuracy, however, degrades as the number of tools grows, since every tool description adds context~\cite{Zhang2026-pd,Paramanayakam2025-av}; most methods therefore either add a procedure that \replaced{filters out which descriptions to present}{selects which descriptions to present}, or start from a small set of tools. Long contexts of this kind are also expensive in energy and compute on the device~\cite{Chen2024-zs}.

\replaced{These methods are anyway}{In any case, these methods are} concerned with a single step, for which they need an external planner~\cite{Chen2024-zs}. Some approaches instead target complete agentic tasks. TinyAgent brings the DAG-based planning approach above to edge devices, showing that finetuned SLMs between 1 and 7B parameters can compete with vanilla GPT-4-Turbo~\cite{Erdogan2024-bh}; the price is a fixed and very limited set of tools (16 in the paper), a curated finetuning dataset which is expensive to build and must be regenerated when new tools appear, an additional procedure to select which function definitions to provide in the prompt, and no recovery mechanism when the initial plan is wrong. \citet{Yang2026-qs} instead combine three SLMs with different roles to carry out agentic tasks in IIoT settings.

What these results show is not that SLMs are unusable, but that they suffer as the context grows, for instance when selecting a function among a few dozen descriptions. Given issues in such a narrow regime, and given that the methods carrying out complete agentic tasks with SLMs succeed only under heavy scaffolding -- a small fixed tool set, a curated finetuning dataset, or an explicit tool-filtering stage -- a natural conclusion is that \emph{at present} SLMs are not appropriate for driving general agentic tasks on their own. A single question asked over a single, bounded piece of data does not enter this regime, and this is precisely the one in which small models are known to be reliable~\cite{Narayan2025-xp,Belcak2025-uc} -- a property we exploit.

\paragraph{Local-remote agent collaboration.}
Since neither placing the whole loop in the cloud nor running it on the device is satisfactory, some systems split the work between the two. In MinionS a local SLM extracts information from documents as instructed by a remote LLM, which aggregates the results~\cite{Narayan2025-xp}. In EcoAgent the LLM generates a sequential plan specifying input and output conditions for every step, and a local SLM carries the steps out and verifies that the progression adheres to the expectation~\cite{Yi2026-wo}. In LEI the edge device periodically uploads sensor readings to the cloud, where the LLM synthesizes the code that is then used to analyze the data directly on the device~\cite{Dehury2026-ye}. In other settings the loop stays in the cloud and a local model sits in front of it, sanitizing the data by replacing sensitive information with placeholders~\cite{Yu2026-ow,Yuan2026-kj} or by re-writing the prompt~\cite{Siyan2024-zf} before uploading.

What is missing is the intersection of these lines: a division that keeps the \replaced{}{cloud} LM out of the loop during execution, as code-first agents do, while still allowing the program to reason over data that only exists at run time. The design we evaluate differs from the above along several axes at once: it studies an explicitly asymmetric placement, with a large planner in the cloud and a small executor on the device; it recovers from errors by regenerating the program from the traceback rather than by re-planning blindly; it makes no domain-specific assumptions, unlike collaborative systems studied so far in narrow domains such as data analysis or sensing; and it employs the SLM for the kind of question small models already answer reliably, rather than introducing finetuning or tool-filtering machinery to make them fit a role they are not suited for. To our knowledge, no existing method combines these properties, and no prior work has quantified how often code-first agents fail simply because the program cannot interpret the data it retrieves.

%% file: appendix/examples.tex
\section{Examples of the \texttt{slm} function usage}
\label{app:examples}

To illustrate the kind of task the proposed architecture targets, we take two examples from the AppWorld benchmark. Both are cases in which a task requiring tens of steps can be carried out almost entirely by pure Python, with an LM being strictly necessary at one or two points only. This is not meant as evidence that the approach works in general, which we quantify in Section~\ref{sec:evaluation}, but as an existence proof of the pattern the method exploits.\footnote{These are not made-up examples, but actual programs we obtained by running the proposed method. We only introduced comments in the listings for clarity.} In both examples, API keys are necessary to access applications, read local files, etc., all of which are by nature private~\cite{Yu2026-ow}.

\begin{example}\label{ex:messages}
Kristin paid for my grocery recently as my payment cards were not working at the time. Send them the owed money with a description note ``Groceries'' as per my phone text conversation, and then send them a phone text message, ``It is done.''
\end{example}

\begin{wrapfigure}[6]{r}{0.72\textwidth}
\vspace{-1.45\intextsep}
\begin{minted}[frame=single, fontsize=\scriptsize,style=perldoc]{python}
amount_owed = slm(
    "What is the dollar amount that Kristin said I owe her for groceries?",
    context=messages, # fetched from the messages API
    context_description="all text messages between me and Kristin",
    keys=["message"], # extract bodies, remove metadata
    output_type="float",
)
\end{minted}
\end{wrapfigure}

Of the 17 steps the emitted program performs, all but one are deterministic: resolving the contact, fetching the conversation, issuing the payment, and sending the confirmation message are all plain API calls with known arguments. The exception is recovering how much is owed, which is stated somewhere in the conversation in natural language. Extracting it with a regular expression is unreliable, since a regex cannot tell the amount apart from a look-alike distractor elsewhere in the thread. An LM is thus needed -- but note \emph{which} one: answering this question requires reading the entire recent message history, so delegating it to the cloud means uploading the conversation in order to obtain a single number. With our design the program instead calls \verb|slm| at that one point.

\begin{example}\label{ex:file-system}
I paid for our last month's electricity bill. Its amount is supposed to be shared equally among my roommates and me. Make venmo requests to my roommates, with a description note, ``For electricity bill.''. The bill receipt is in my file system.
\end{example}

\begin{wrapfigure}[5]{r}{0.64\textwidth}
\vspace{-1.3 \intextsep}
\begin{minted}[frame=single, fontsize=\scriptsize,style=perldoc]{python}
electricity_bill_path = slm(
    "Which of these files is the electricity bill for last month?",
    context=files, # survivors of a keyword filter
    output_type="item", # return one element of files
)
\end{minted}
\end{wrapfigure}

In the second example, the user never says which file contains the receipt, and the name is unknown at generation time -- it may be \verb|May23.txt|, \verb|May_23.txt|, \verb|05_2023.txt|, or anything else, anywhere in the file system. Left to guess, the LLM hallucinates a plausible path, which is what we observe in practice. The division of labor we advocate resolves this cheaply: pure Python walks the file system and keeps the files whose name contains ``electric'', ``bill'', or a few related terms, and the SLM is asked a single question over the handful of survivors. Guessing a keyword set is recoverable in a way that guessing a path is not, since the filter is deliberately high-recall and its failures surface as a wrong or empty candidate set rather than as a non-existent path.

\begin{wrapfigure}[5]{r}{0.55\textwidth}
\vspace{-1.3\intextsep}
\begin{minted}[frame=single, fontsize=\scriptsize,style=perldoc]{python}
total_amount = slm(
    "What is the total amount of the electricity bill?",
    context=bill["content"],
    output_type="float",
)
\end{minted}
\end{wrapfigure}
Having identified the file containing the receipt, the program must extract the amount that was paid. As in \Cref{ex:messages}, a regular expression is not a reliable option, and for the same underlying reason: the program is written before any file is read, so its layout is unknown at generation time. A receipt may list several currency values, such as an amount before taxes, a tax line, and a total, and there is no guarantee that the figure of interest carries a dollar sign at all, as in ``Total (\$): 122.30''. Distinguishing the total from the other values requires reading the labels around it, which is exactly the kind of interpretation a pattern fixed in advance cannot perform. The generated program therefore delegates this step to the SLM as well.

%% file: appendix/reproducibility.tex
\section{Reproducibility details}\label{app:reproducibility}

We serve every model locally with \texttt{llama.cpp}, compiled from source (build \texttt{b10162}, commit \texttt{6ba5ef247}) separately for the A100 and V100 GPUs. \Cref{tab:repro-models} lists the checkpoints and the hardware. Each model runs on a single GPU, without tensor parallelism.

\begin{table}[h]
\centering\footnotesize
\caption{Employed models. All weights are GGUF files (UD = Unsloth Dynamic quantization).}
\label{tab:repro-models}
\begin{tabular}{@{}lll@{}}
\toprule
Model & Quantization & GPU \\
\midrule
Qwen-27B (Qwen3.6-27B) & UD-Q8\_K\_XL & A100-80GB \\
Qwen-122B (Qwen3.5-122B-A10B) & UD-Q4\_K\_XL & A100-80GB \\
Qwen-coder (Qwen3-Coder-Next) & Q6\_K & A100-80GB \\
Qwen3.5-0.8B/2B/4B/9B & UD-Q8\_K\_XL & V100 \\
Gemma-4 E2B--31B & UD-Q8\_K\_XL & A100-80GB \\
\bottomrule
\end{tabular}
\end{table}

\paragraph{Generation.}
Decoding is greedy: every request uses temperature 0, and we set neither top-$p$ nor top-$k$.
Qwen-27B and Qwen-122B models produce a thinking trace before answering. Qwen-coder produces none, and Gemma-4 runs with thinking disabled, its chat-template default.
To stop runaway generations, planner and ReAct outputs are capped at 16,384 tokens, thinking included. For the SLM we cap thinking itself at 4,096 tokens; the server then appends a short wind-down message and closes the thinking block, so the model still answers.

\paragraph{Protocol.}
We evaluate on the 168 Test Normal and 417 Test Challenge tasks of AppWorld (data v0.2.0) and report success rate (referred to as ``task goal completion'' in the original repository). The evaluator is unchanged from upstream commit a072b7a, which our fork is based on.
We run each configuration once. Standard errors come from bootstrapping over tasks: 10,000 resamples, with the SE taken as the standard deviation of the resampled means.

\paragraph{Baseline Agents.}
FC is AppWorld's full-code agent. Besides minor fixes, we extend it in two ways, and our agent inherits both. 
First, a single planner call, prompted with three training tasks as demonstrations, shortlists up to 20 APIs. The list is completed with the login and ID-provider APIs it depends on, and it is cached, so FC and our agent receive the same shortlist for a given planner.
Second, the planner may look up API documentation at run time through documentation-only programs, which run on the server, as the documentation contains no user data. The first four such programs of a task do not count as attempts, although each one is a planner call.
Each agent then makes at most five attempts, i.e., generates at most five programs that are executed on the device; the shortlisting call above runs entirely in the cloud and does not count as an attempt. After every attempt that raises an error, the planner receives the untruncated traceback, together with AppWorld's reflexion prompt, and the whole conversation is kept.
The ReAct baseline is AppWorld's ReAct agent, unmodified. It runs for at most 50 steps and never truncates observations or history.

\paragraph{SLM answer quality study.}
For the agreement study, we take every 9B round-trip recorded in the three TN runs: 1,584 in total (676 \texttt{bool}, 663 \texttt{str}, 209 \texttt{item} and 36 numeric). We replay each one on every candidate SLM with the same prompt and decoding settings.
Two answers agree when their outcomes match and, if both succeed, their cast values are equal. Floats must match exactly. Strings are compared after removing the thinking trace, canonicalizing JSON and collapsing whitespace.
Error bars are 95\% percentile intervals over 1,000 bootstrap resamples of tasks.

%% file: main.bbl
\begin{thebibliography}{45}
\providecommand{\natexlab}[1]{#1}
\providecommand{\url}[1]{\texttt{#1}}
\expandafter\ifx\csname urlstyle\endcsname\relax
  \providecommand{\doi}[1]{doi: #1}\else
  \providecommand{\doi}{doi: \begingroup \urlstyle{rm}\Url}\fi

\bibitem[Belcak et~al.(2025)Belcak, Heinrich, Diao, Fu, Dong, Muralidharan, Lin, and Molchanov]{Belcak2025-uc}
Peter Belcak, Greg Heinrich, Shizhe Diao, Yonggan Fu, Xin Dong, Saurav Muralidharan, Yingyan~Celine Lin, and Pavlo Molchanov.
\newblock {Small Language Models are the Future of Agentic {AI}}, 2025.
\newblock arXiv:\href{https://arxiv.org/abs/2506.02153}{2506.02153}.

\bibitem[Cai et~al.(2024)Cai, Wang, Ma, Chen, and Zhou]{cai2024large}
Tianle Cai, Xuezhi Wang, Tengyu Ma, Xinyun Chen, and Denny Zhou.
\newblock \href{https://proceedings.iclr.cc/paper_files/paper/2024/hash/ed91353f700d113e5d848c7e04a858b0-Abstract-Conference.html}{Large language models as tool makers}.
\newblock In \emph{ICLR}, 2024.

\bibitem[Chen and Li(2024)]{Chen2024-hf}
Wei Chen and Zhiyuan Li.
\newblock {Octopus v2: On-device Language Model for Super Agent}.
\newblock 2024.
\newblock arXiv:\href{https://arxiv.org/abs/2404.01744}{2404.01744}.

\bibitem[Chen et~al.(2025{\natexlab{a}})Chen, Li, Guo, and Shen]{Chen2024-zs}
Wei Chen, Zhiyuan Li, Zhen Guo, and Yikang Shen.
\newblock {Octo-Planner: On-Device Language Model for Planner-Action Agents}.
\newblock In \emph{EMAS}, 2025{\natexlab{a}}.
\newblock \doi{10.1007/978-3-032-18011-7\_9}.

\bibitem[Chen et~al.(2025{\natexlab{b}})Chen, Li, and Ma]{Chen2024-sa}
Wei Chen, Zhiyuan Li, and Mingyuan Ma.
\newblock {Octopus: On-device language model for function calling of software {APIs}}.
\newblock In \emph{NAACL}, 2025{\natexlab{b}}.
\newblock \doi{10.18653/V1/2025.NAACL-INDUSTRY.27}.

\bibitem[Chen et~al.(2023)Chen, Ma, Wang, and Cohen]{Chen2022-mh}
Wenhu Chen, Xueguang Ma, Xinyi Wang, and William~W. Cohen.
\newblock \href{https://openreview.net/forum?id=YfZ4ZPt8zd}{Program of Thoughts Prompting: Disentangling Computation from Reasoning for Numerical Reasoning Tasks}.
\newblock \emph{Trans. Mach. Learn. Res.}, 2023.

\bibitem[Debenedetti et~al.(2026)Debenedetti, Shumailov, Fan, Hayes, Carlini, Fabian, Kern, Shi, Terzis, and Tram{\`e}r]{debenedetti2026defeating}
Edoardo Debenedetti, Ilia Shumailov, Tianqi Fan, Jamie Hayes, Nicholas Carlini, Daniel Fabian, Christoph Kern, Chongyang Shi, Andreas Terzis, and Florian Tram{\`e}r.
\newblock Defeating prompt injections by design.
\newblock In \emph{SaTML}, 2026.
\newblock \doi{10.1109/SaTML68715.2026.00040}.

\bibitem[Dehury et~al.(2026)Dehury, Kushwaha, Zhang, Saleh, and Donta]{Dehury2026-ye}
Chinmaya~Kumar Dehury, Siddharth~Singh Kushwaha, Qiyang Zhang, Alaa Saleh, and Praveen~Kumar Donta.
\newblock {LLM-assisted Agentic Edge Intelligence Framework}.
\newblock 2026.
\newblock arXiv:\href{https://arxiv.org/abs/2604.09607}{2604.09607}.

\bibitem[Erdogan et~al.(2024)Erdogan, Lee, Jha, Kim, Tabrizi, Moon, Hooper, Anumanchipalli, Keutzer, and Gholami]{Erdogan2024-bh}
Lutfi~Eren Erdogan, Nicholas Lee, Siddharth Jha, Sehoon Kim, Ryan Tabrizi, Suhong Moon, Coleman Richard~Charles Hooper, Gopala Anumanchipalli, Kurt Keutzer, and Amir Gholami.
\newblock {{TinyAgent}: Function Calling at the Edge}.
\newblock In \emph{EMNLP}, 2024.
\newblock \doi{10.18653/V1/2024.EMNLP-DEMO.9}.

\bibitem[Erdogan et~al.(2025)Erdogan, Lee, Kim, Moon, Furuta, Anumanchipalli, Keutzer, and Gholami]{Erdogan2025-xr}
Lutfi~Eren Erdogan, Nicholas Lee, Sehoon Kim, Suhong Moon, Hiroki Furuta, Gopala Anumanchipalli, Kurt Keutzer, and Amir Gholami.
\newblock \href{https://openreview.net/forum?id=ybA4EcMmUZ}{Plan-and-Act: Improving Planning of Agents for Long-Horizon Tasks}.
\newblock In \emph{ICML}, 2025.

\bibitem[Gao et~al.(2023)Gao, Madaan, Zhou, Alon, Liu, Yang, Callan, and Neubig]{Gao2022-mj}
Luyu Gao, Aman Madaan, Shuyan Zhou, Uri Alon, Pengfei Liu, Yiming Yang, Jamie Callan, and Graham Neubig.
\newblock \href{https://proceedings.mlr.press/v202/gao23f.html}{{PAL:} Program-aided Language Models}.
\newblock In \emph{ICML}, 2023.

\bibitem[Giusti et~al.(2025)Giusti, Werner, Taiello, Costa, Tosun, Protani, Molina, de~Almeida, Cacace, Santos, and Serio]{Giusti2025-yo}
Lorenzo Giusti, Ole~Anton Werner, Riccardo Taiello, Matilde~Carvalho Costa, Emre Tosun, Andrea Protani, Marc Molina, Rodrigo~Lopes de~Almeida, Paolo Cacace, Diogo~Reis Santos, and Luigi Serio.
\newblock \href{https://openreview.net/forum?id=N7NDfV2YMp}{Federation of Agents: {A} Semantics-Aware Communication Fabric for Large-Scale Agentic {AI}}.
\newblock In \emph{ER (NeurIPS Workshop)}, 2025.

\bibitem[Gupta and Kembhavi(2023)]{Gupta2022-ac}
Tanmay Gupta and Aniruddha Kembhavi.
\newblock {Visual Programming: Compositional Visual Reasoning Without Training}.
\newblock In \emph{CVPR}, 2023.
\newblock \doi{10.1109/CVPR52729.2023.01436}.

\bibitem[Haque et~al.(2025)Haque, Rahman, Gupta, Shujaee, and George]{Haque2025-bq}
Mohd~Ariful Haque, Fahad Rahman, Kishor~Datta Gupta, Khalil Shujaee, and Roy George.
\newblock {TinyLLM: Evaluation and Optimization of Small Language Models for Agentic Tasks on Edge Devices}.
\newblock 2025.
\newblock arXiv:\href{https://arxiv.org/abs/2511.22138}{2511.22138}.

\bibitem[Harikumar(2026)]{Harikumar2026-rz}
Pranav Harikumar.
\newblock {PlanCompiler: A Deterministic Compilation Architecture for Structured Multi-Step LLM Pipelines}.
\newblock 2026.
\newblock \doi{10.21203/rs.3.rs-9186427/v1}.

\bibitem[Hou et~al.(2024)Hou, Yang, Jiao, Wang, Tu, and Zhao]{Hou2024-fz}
Xinming Hou, Mingming Yang, Wenxiang Jiao, Xing Wang, Zhaopeng Tu, and Wayne~Xin Zhao.
\newblock {CoAct: {A} Global-Local Hierarchy for Autonomous Agent Collaboration}.
\newblock 2024.
\newblock arXiv:\href{https://arxiv.org/abs/2406.13381}{2406.13381}.

\bibitem[Kadekodi et~al.(2025)Kadekodi, Jin, Kamahori, Gu, Khatiri, Bayindirli, Gorbunov, and Kasikci]{Kadekodi2025-sq}
Rohan Kadekodi, Zhan Jin, Keisuke Kamahori, Yile Gu, Sean Khatiri, Noah~H Bayindirli, Sergey Gorbunov, and Baris Kasikci.
\newblock {{AgentFlux}}: Decoupled fine-tuning \& inference for on-device agentic systems.
\newblock 2025.
\newblock arXiv:\href{https://arxiv.org/abs/2510.00229}{2510.00229}.

\bibitem[Kaplan et~al.(2020)Kaplan, McCandlish, Henighan, Brown, Chess, Child, Gray, Radford, Wu, and Amodei]{Kaplan2020-es}
Jared Kaplan, Sam McCandlish, Tom Henighan, Tom~B. Brown, Benjamin Chess, Rewon Child, Scott Gray, Alec Radford, Jeffrey Wu, and Dario Amodei.
\newblock {Scaling Laws for Neural Language Models}.
\newblock 2020.
\newblock arXiv:\href{https://arxiv.org/abs/2001.08361}{2001.08361}.

\bibitem[Kim et~al.(2024)Kim, Moon, Tabrizi, Lee, Mahoney, Keutzer, and Gholami]{Kim2023-rh}
Sehoon Kim, Suhong Moon, Ryan Tabrizi, Nicholas Lee, Michael~W. Mahoney, Kurt Keutzer, and Amir Gholami.
\newblock \href{https://proceedings.mlr.press/v235/kim24y.html}{An LLM Compiler for Parallel Function Calling}.
\newblock In \emph{ICML}, 2024.

\bibitem[Li et~al.(2025)Li, Raghuram, Khattab, Hirschberg, and Yu]{Siyan2024-zf}
Siyan Li, Vethavikashini~Chithrra Raghuram, Omar Khattab, Julia Hirschberg, and Zhou Yu.
\newblock {{PAPILLON:} Privacy Preservation from Internet-based and Local Language Model Ensembles}.
\newblock In \emph{NAACL}, 2025.
\newblock \doi{10.18653/V1/2025.NAACL-LONG.173}.

\bibitem[Lin et~al.(2024)Lin, Wen, Peng, Nie, Liao, Wang, Mo, Zhou, Cheng, Zhao, Wang, and Zhang]{Lin2024-ce}
Qiqiang Lin, Muning Wen, Qiuying Peng, Guanyu Nie, Junwei Liao, Jun Wang, Xiaoyun Mo, Jiamu Zhou, Cheng Cheng, Yin Zhao, Jun Wang, and Weinan Zhang.
\newblock {Hammer: Robust Function-Calling for On-Device Language Models via Function Masking}.
\newblock 2024.
\newblock arXiv:\href{https://arxiv.org/abs/2410.04587}{2410.04587}.

\bibitem[Narayan et~al.(2025)Narayan, Biderman, Eyuboglu, May, Linderman, Zou, and R{\'{e}}]{Narayan2025-xp}
Avanika Narayan, Dan Biderman, Sabri Eyuboglu, Avner May, Scott~W. Linderman, James Zou, and Christopher R{\'{e}}.
\newblock \href{https://openreview.net/forum?id=qGDlzt3dKz}{{Minions: Cost-efficient Collaboration Between On-device and Cloud Language Models}}.
\newblock In \emph{ICML}, 2025.

\bibitem[Paprunia et~al.(2025)Paprunia, Kharidia, and Doshi]{Paprunia2025-cm}
Dhruvi Paprunia, Vansh Kharidia, and Pankti Doshi.
\newblock {Advancing SLM Tool-Use Capability using Reinforcement Learning}.
\newblock In \emph{AIC}, 2025.
\newblock \doi{10.1109/AIC66080.2025.11211997}.

\bibitem[Paramanayakam et~al.(2025)Paramanayakam, Karatzas, Anagnostopoulos, and Stamoulis]{Paramanayakam2025-av}
Varatheepan Paramanayakam, Andreas Karatzas, Iraklis Anagnostopoulos, and Dimitrios Stamoulis.
\newblock {Less is More: Optimizing Function Calling for {LLM} Execution on Edge Devices}.
\newblock In \emph{DATE}, 2025.
\newblock \doi{10.23919/DATE64628.2025.10992798}.

\bibitem[Prasad et~al.(2024)Prasad, Koller, Hartmann, Clark, Sabharwal, Bansal, and Khot]{Prasad2023-sv}
Archiki Prasad, Alexander Koller, Mareike Hartmann, Peter Clark, Ashish Sabharwal, Mohit Bansal, and Tushar Khot.
\newblock {{ADaPT}: As-Needed Decomposition and Planning with Language Models}.
\newblock In \emph{Findings of NAACL}, 2024.
\newblock \doi{10.18653/V1/2024.FINDINGS-NAACL.264}.

\bibitem[Rainone et~al.(2026)Rainone, Belli, Major, and Behboodi]{Rainone2026-op}
Corrado Rainone, Davide Belli, Bence Major, and Arash Behboodi.
\newblock {When Cloud Agents Meet Device Agents: Lessons from Hybrid Multi-Agent Systems}.
\newblock 2026.
\newblock arXiv:\href{https://arxiv.org/abs/2605.30102}{2605.30102}.

\bibitem[Schick et~al.(2023)Schick, Dwivedi{-}Yu, Dess{\`{\i}}, Raileanu, Lomeli, Hambro, Zettlemoyer, Cancedda, and Scialom]{Schick2023-fy}
Timo Schick, Jane Dwivedi{-}Yu, Roberto Dess{\`{\i}}, Roberta Raileanu, Maria Lomeli, Eric Hambro, Luke Zettlemoyer, Nicola Cancedda, and Thomas Scialom.
\newblock \href{http://papers.nips.cc/paper\_files/paper/2023/hash/d842425e4bf79ba039352da0f658a906-Abstract-Conference.html}{Toolformer: Language Models Can Teach Themselves to Use Tools}.
\newblock In \emph{NeurIPS}, 2023.

\bibitem[Shen et~al.(2023)Shen, Song, Tan, Li, Lu, and Zhuang]{Shen2023-oi}
Yongliang Shen, Kaitao Song, Xu~Tan, Dongsheng Li, Weiming Lu, and Yueting Zhuang.
\newblock \href{http://papers.nips.cc/paper\_files/paper/2023/hash/77c33e6a367922d003ff102ffb92b658-Abstract-Conference.html}{{HuggingGPT}: Solving {AI} Tasks with {ChatGPT} and its Friends in Hugging Face}.
\newblock In \emph{NeurIPS}, 2023.

\bibitem[Sun et~al.(2026)Sun, Zhu, Xu, Du, Li, and Lo]{sun2026toward}
Zhensu Sun, Haotian Zhu, Bowen Xu, Xiaoning Du, Li~Li, and David Lo.
\newblock Toward agentic runtime healing.
\newblock \emph{Commun. ACM}, 2026.
\newblock \doi{10.1145/3800691}.

\bibitem[Surís et~al.(2023)Surís, Menon, and Vondrick]{Suris2023-yy}
Dídac Surís, Sachit Menon, and Carl Vondrick.
\newblock {{ViperGPT}: Visual Inference via Python Execution for Reasoning}.
\newblock In \emph{ICCV}, 2023.
\newblock \doi{10.1109/ICCV51070.2023.01092}.

\bibitem[Trivedi et~al.(2024)Trivedi, Khot, Hartmann, Manku, Dong, Li, Gupta, Sabharwal, and Balasubramanian]{Trivedi2024-qn}
Harsh Trivedi, Tushar Khot, Mareike Hartmann, Ruskin Manku, Vinty Dong, Edward Li, Shashank Gupta, Ashish Sabharwal, and Niranjan Balasubramanian.
\newblock {{AppWorld}: {A} Controllable World of Apps and People for Benchmarking Interactive Coding Agents}.
\newblock In \emph{ACL}, 2024.
\newblock \doi{10.18653/V1/2024.ACL-LONG.850}.

\bibitem[Wang et~al.(2024)Wang, Chen, Yuan, Zhang, Li, Peng, and Ji]{Wang2024-rm}
Xingyao Wang, Yangyi Chen, Lifan Yuan, Yizhe Zhang, Yunzhu Li, Hao Peng, and Heng Ji.
\newblock \href{https://proceedings.mlr.press/v235/wang24h.html}{Executable Code Actions Elicit Better {LLM} Agents}.
\newblock In \emph{ICML}, 2024.

\bibitem[Wang et~al.(2023)Wang, Cai, Liu, Ma, and Liang]{Wang2023-cx}
Zihao Wang, Shaofei Cai, Anji Liu, Xiaojian Ma, and Yitao Liang.
\newblock \href{https://proceedings.neurips.cc/paper_files/paper/2023/hash/6b8dfb8c0c12e6fafc6c256cb08a5ca7-Abstract-Conference.html}{Describe, Explain, Plan and Select: Interactive Planning with Large Language Models Enables Open-World Multi-Task Agents}.
\newblock In \emph{NeurIPS}, 2023.

\bibitem[Wei et~al.(2026)Wei, Dong, Wang, Zhang, Zhao, Shen, Xia, and Yin]{Wei2025-wx}
Xiaolong Wei, Yuehu Dong, Xingliang Wang, Xingyu Zhang, Zhejun Zhao, Dongdong Shen, Long Xia, and Dawei Yin.
\newblock {Beyond {ReAct}: {A} Planner-Centric Framework for Complex Tool-Augmented {LLM} Reasoning}.
\newblock In \emph{AAAI}, 2026.
\newblock \doi{10.1609/AAAI.V40I40.40676}.

\bibitem[Wen et~al.(2025)Wen, Tian, Pavlov, Du, Li, Chang, Zhao, Liu, Liu, Zhang, and Li]{Wen2025-as}
Hao Wen, Shizuo Tian, Borislav Pavlov, Wenjie Du, Yixuan Li, Ge~Chang, Shanhui Zhao, Jiacheng Liu, Yunxin Liu, Ya-Qin Zhang, and Yuanchun Li.
\newblock {{AutoDroid-V2}: Boosting {SLM-based} {GUI} Agents via Code Generation}.
\newblock In \emph{MobiSys}, 2025.
\newblock \doi{10.1145/3711875.3729134}.

\bibitem[Wen and Chen(2026)]{Yihan2026-ya}
Yihan Wen and Xin Chen.
\newblock {PseudoAct: Leveraging Pseudocode Synthesis for Flexible Planning and Action Control in Large Language Model Agents}.
\newblock 2026.
\newblock arXiv:\href{https://arxiv.org/abs/2602.23668}{2602.23668}.

\bibitem[Willison(2023)]{willison2023dual}
Simon Willison.
\newblock The {Dual LLM} pattern for building {AI} assistants that can resist prompt injection.
\newblock \url{https://simonwillison.net/2023/Apr/25/dual-llm-pattern/}, 2023.

\bibitem[Xu et~al.(2023)Xu, Peng, Lei, Mukherjee, Liu, and Xu]{Xu2023-xt}
Binfeng Xu, Zhiyuan Peng, Bowen Lei, Subhabrata Mukherjee, Yuchen Liu, and Dongkuan Xu.
\newblock {ReWOO: Decoupling Reasoning from Observations for Efficient Augmented Language Models}.
\newblock 2023.
\newblock arXiv:\href{https://arxiv.org/abs/2305.18323}{2305.18323}.

\bibitem[Yang et~al.(2026)Yang, Li, Tang, Liu, Guo, and Wen]{Yang2026-qs}
Qingwen Yang, Zhi Li, Jiawei Tang, Yanyi Liu, Tiezheng Guo, and Yingyou Wen.
\newblock {{LEAF}}: A lightweight edge agent framework with expert {SLMs} for the industrial internet of things.
\newblock \emph{Comput. Mater. Contin.}, 2026.
\newblock \doi{10.32604/cmc.2025.074384}.

\bibitem[Yao et~al.(2023)Yao, Zhao, Yu, Du, Shafran, Narasimhan, and Cao]{Yao2022-pw}
Shunyu Yao, Jeffrey Zhao, Dian Yu, Nan Du, Izhak Shafran, Karthik~R. Narasimhan, and Yuan Cao.
\newblock \href{https://openreview.net/forum?id=WE\_vluYUL-X}{{ReAct: Synergizing Reasoning and Acting in Language Models}}.
\newblock In \emph{ICLR}, 2023.

\bibitem[Yi et~al.(2026)Yi, Hu, Chen, Zhang, Yang, and Wu]{Yi2026-wo}
Biao Yi, Xueyu Hu, Yurun Chen, Shengyu Zhang, Hongxia Yang, and Fan Wu.
\newblock {{EcoAgent}: An Efficient Device-Cloud Collaborative Multi-Agent Framework for Mobile Automation}.
\newblock In \emph{AAAI}, 2026.
\newblock \doi{10.1609/AAAI.V40I35.40230}.

\bibitem[Yu et~al.(2026)Yu, Wang, Lang, Su, and Wang]{Yu2026-ow}
Guangsheng Yu, Qin Wang, Rui Lang, Shuai Su, and Xu~Wang.
\newblock {{PlanTwin}: Privacy-Preserving Planning Abstractions for Cloud-Assisted {LLM} Agents}.
\newblock \emph{{IEEE} Trans. Serv. Comput.}, 2026.
\newblock \doi{10.1109/TSC.2026.3695494}.

\bibitem[Yuan et~al.(2026)Yuan, Fang, Wang, and Brinton]{Yuan2026-kj}
Liangqi Yuan, Wenzhi Fang, Shiqiang Wang, and Christopher~G. Brinton.
\newblock {PAAC: Privacy-Aware Agentic Device-Cloud Collaboration}.
\newblock 2026.
\newblock arXiv:\href{https://arxiv.org/abs/2605.08646}{2605.08646}.

\bibitem[Zhang et~al.(2025)Zhang, Wornow, and Olukotun]{Zhang2026-kp}
Qizheng Zhang, Michael Wornow, and Kunle Olukotun.
\newblock \href{http://papers.nips.cc/paper\_files/paper/2025/hash/9549f7d06700f0966d5f938f1d11022a-Abstract-Conference.html}{Agentic Plan Caching: Test-Time Memory for Fast and Cost-Efficient {LLM} Agents}.
\newblock In \emph{NeurIPS}, 2025.

\bibitem[Zhang et~al.(2026)Zhang, Wang, He, Wang, and Li]{Zhang2026-pd}
Xiaoyu Zhang, Yi~Wang, Along He, Haobin Wang, and Tao Li.
\newblock {{TAPE:} {A} Multi-Agent Framework for Task-Adaptive Planning and Execution in Resource-Constrained Environments}.
\newblock \emph{Expert Syst. Appl.}, 2026.
\newblock \doi{10.1016/J.ESWA.2025.129423}.

\end{thebibliography}
